\documentclass[letterpaper, 10 pt, conference]{ieeeconf}  
\pdfoutput=1 

\IEEEoverridecommandlockouts                              

\usepackage{graphicx} 
\usepackage{amsmath} 
\usepackage{amssymb}  
\usepackage{todonotes}
\usepackage{xcolor}
\usepackage{subfig}
\usepackage[T1]{fontenc}
\usepackage[font=small]{caption}
\usepackage{booktabs}
\usepackage{url}
\usepackage[margin=.8in]{geometry}

\newcommand{\etal}{\textit{et al. }}

\renewcommand{\L}{\mathcal{L}}
\DeclareMathOperator{\E}{\mathbb{E}}

\title{\LARGE \bf
Learning to Drive from Simulation without Real World Labels
}

\author{Alex Bewley, Jessica Rigley, Yuxuan Liu, Jeffrey Hawke, Richard Shen, Vinh-Dieu Lam, Alex Kendall
\thanks{The authors are with Wayve in Cambridge, UK.}
\thanks{{\tt\small research@wayve.ai}}%
}

\begin{document}

\maketitle
\thispagestyle{empty}
\pagestyle{empty} %

\begin{abstract}

Simulation can be a powerful tool for understanding machine learning systems and designing methods to solve real-world problems. Training and evaluating methods purely in simulation is often ``doomed to succeed'' at the desired task in a simulated environment, but the resulting models are incapable of operation in the real world. Here we present and evaluate a method for transferring a vision-based lane following driving policy from simulation to operation on a rural road without any real-world labels. Our approach leverages recent advances in image-to-image translation to achieve domain transfer while jointly learning a single-camera control policy from simulation control labels. We assess the driving performance of this method using both open-loop regression metrics, and closed-loop performance operating an autonomous vehicle on rural and urban roads.

\end{abstract}



\section{Introduction}


This paper demonstrates how to use machine translation techniques for unsupervised transfer of an end-to-end driving model from a simulated environment to a real vehicle. We trained a deep learning model to drive in a simulated environment (where complete knowledge of the environment is possible) and adapted it for the visual variation experienced in the real world (completely unsupervised and without real-world labels). This work goes beyond simple image-to-image translation by making the desired task of driving a differentiable component within a deep learning architecture. We show this shapes the learning process to ensure the driving policy is invariant to the visual variation across both simulated and real domains. 

Learning to drive in simulation offers a number of benefits, including the ability to vary appearance, lighting and weather conditions along with more structural variations, such as curvature of the road and complexity of the surrounding obstructions. It is also possible to construct scenarios in simulation which are difficult (or dangerous) to replicate in the real world. 
Furthermore, simulation provides ground-truth representations like semantics and even privileged information such as the relative angle and position of the vehicle with respect to the road \cite{richter2016playing}. However, applying this knowledge to real-world applications has been limited due to the reality gap, often expressed as the difference in appearance from what can be rendered in simulation to how the physical world is viewed from a camera. 

In contrast to the many efforts to improve the photometric realism of rendered images \cite{Dosovitskiy17, gaidon2016virtual,richter2016playing,ros2016synthia}, this work learns an invariant mapping between features observed in simulation and the real environment. This representation is jointly optimised for steering a vehicle along a road in addition to image translation. While enjoying the benefits of simulation, this method is also adept at inferring the common structure between the target environment and what has been seen in simulation. This is analogous to transferring skills learned though playing virtual racing games to a person's first experience behind the wheel of a car \cite{li2016playing}. 


\begin{figure}[!t]
	\centering
	\includegraphics[width=\columnwidth, scale=1, bb=0 0 407 358]{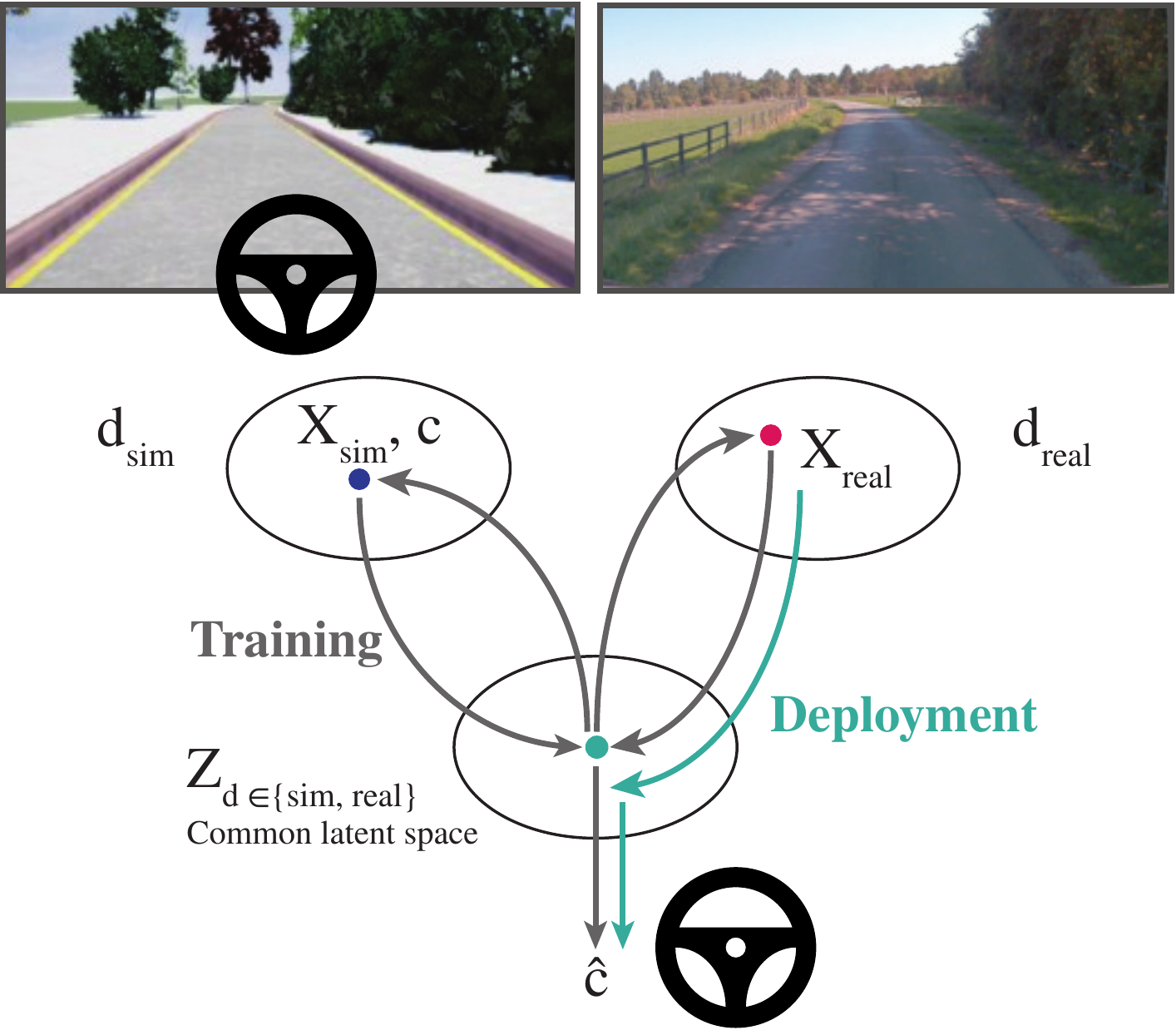}
	\caption{We constructed a model for end-to-end driving (vision to action) for lane following by learning to translate between simulated and real-world imagery ($X_{sim,real}$ from domains $d_{sim,real}$), jointly learning a control policy from this common latent space $Z$ using labels $c$ from an expert driver in simulation. This method does not require real-world control labels (which are more difficult to acquire), yet learns a policy predicting a control signal $\hat{c}$ which can be applied successfully to a real-world driving task.}
	\label{fig:sim2realcontrolnet-main}
	\vspace{-3mm}
\end{figure}

In summary, the main contributions of this work are:
\begin{itemize}
	\item We present the first example of an end-to-end driving policy transferred from a simulation domain with control labels to an unlabelled real-world domain.
    \item By exploiting the freedom to control the simulated environment we were able to learn manoeuvres in states beyond the common driving distribution in real-world imitation learning, removing the need for multiple camera data collection rigs or data augmentation.
    \item We evaluated this method against a number of baselines using open-loop (offline) and closed-loop metrics, driving a real vehicle over 3km without intervention on a rural road, trained without any real-world labels.
    \item Finally, we demonstrate this method successfully driving in closed-loop on public UK urban roads.
\end{itemize}
A supplementary video demonstrating the system behaviour is available here: \url{https://wayve.ai/sim2real}

\section{Related Work}
There has been significant development in building simulators for the training and evaluation of robotic control policies. While recent simulated environments, such as \cite{richter2017playing, Dosovitskiy17}, have benefited from continued progress in photo realistic gaming engines, the reality gap remains challenging.
\cite{Tobin2017} address this problem by exploiting a simulator's ability to render significant amounts of random colours and textures with the assumption that visual variation observed in the real world is covered.

An active area of research is in image-to-image translation which aims to learn a mapping function between two image distributions \cite{Liu2016a, Liu2017, Isola_2017_CVPR, CycleGAN2017}. 
Isola \etal \cite{Isola_2017_CVPR} proposed a conditional adversarial network where the discriminator compares a pair of images corresponding to two modalities of the same scene.
Zhu \etal \cite{CycleGAN2017} relaxed the need for corresponding images by proposing a cyclic consistency constraint for back translation of images. Similarly,
Liu \etal \cite{Liu2017} combine cross domain discriminators and a cycle consistent reconstruction loss to produce high quality translations. We extend this framework to accommodate learning control as an additional supervised decoder output.

A problem closer to this work is that of unsupervised domain adaptation combining supervision from the source domain with pairwise losses or adversarial techniques to align intermediate features for a supervised task \cite{tzeng2016adapting,wulfmeier2017addressing, Hoffman2018cycada}. However, feature based alignment typically requires some additional pairwise constraints, \cite{tzeng2016adapting} or intermediate domains \cite{wulfmeier2017addressing, wulfmeier2017incremental}, to bridge significant shifts in domain appearance. Other related work align both feature and pixel level distributions for the perception only task of semantic segmentation \cite{Hoffman2018cycada}. The very recent work of Wenzel \etal \cite{wenzel2018modular} learns  transfer control knowledge across simulated weather conditions via a modular pipeline with both image translation and feature-wise alignment.

While the large majority of work in autonomous driving focuses on rule-based and traditional robotic control approaches, end-to-end learning approaches have long held the promise of learning driving policies directly from data with minimal use of brittle assumptions and expensive hand tuning. Pomerleau \cite{pomerleau1989alvinn} first demonstrated the potential feasibility of this approach by successfully applying imitation learning (IL) to a simple real-world lane following task. Muller \etal \cite{muller2006off} further successfully learned to imitate functional driving policies in more unconstrained off-road environments with toy remote control cars. 

More recently, Bojarski \etal \cite{bojarski2016end} demonstrated that simple behaviour cloning could be scaled to a larger and more complex set of lane following scenarios when using a multi-camera system and modern neural net architectures. Codevilla \etal \cite{Codevilla2018} took a similar multi-camera approach, and learned to navigate successfully in simulated urban driving scenarios by conditioning on higher level driving decisions. Building on this, M\"uller \etal \cite{muller2018driving} transferred policies from simulation to reality by using a modular architecture predicting control decisions learned from semantic image segmentation.  

Mehta \etal \cite{mehta2018learning} take a learning from demonstration approach, and augment the control prediction task to depend on intermediate predictions of both high-level visual affordances and low-level action primitives. Kendall \etal \cite{kendall2018learning} demonstrated that pure reinforcement learning approaches with a single camera system could be successfully applied for learning real-world driving policies, though single camera systems have typically not been implemented to date, attributed largely to the robustness gained from training data with multiple views and synthetic control labels. In this work, robustness is achieved through noisy experiences captured in simulation.

In this work, we use an intermediate representation where an implicit latent structure is formed, as opposed to more explicit representations, such as semantic segmentation \cite{hong2018virtual}. 
The work presented in this paper is closest to \cite{Yang2018} who recently proposed to exploit image translation for learning to steer while also sharing a similar philosophy in reducing the complexity of the simulated environment to minimise the sufficient statistics for the desired task. 
Our work differs in that we use a bi-directional image translation model where we are not dependent on having real-world driving commands. We also demonstrate the efficacy of a driving policy on a real system, showing closed-loop performance metrics rather than solely open-loop metrics against a validation set.

\section{Method}
\label{sec:method}

\label{sec:problem}
The problem of translating knowledge from expert demonstrations in simulation to improved real-world behaviour can be cast into the framework of unsupervised domain adaptation. Here, the
training set consisted of image-label pairs from a simulated environment $(X_{sim}, C_{sim})$. An independent set of unlabelled images $(X_{real})$ from the real target environment was used to capture the distribution of real-world observations. Crucially, it is important to note there were no pairwise correspondences between images in the simulated training set and the unlabelled real-world image set.


\subsection{Architecture} 
\label{sec:architecture}

\begin{figure}[t]
\centering
\includegraphics[width=\columnwidth, scale=1, bb=0 0 527 230]{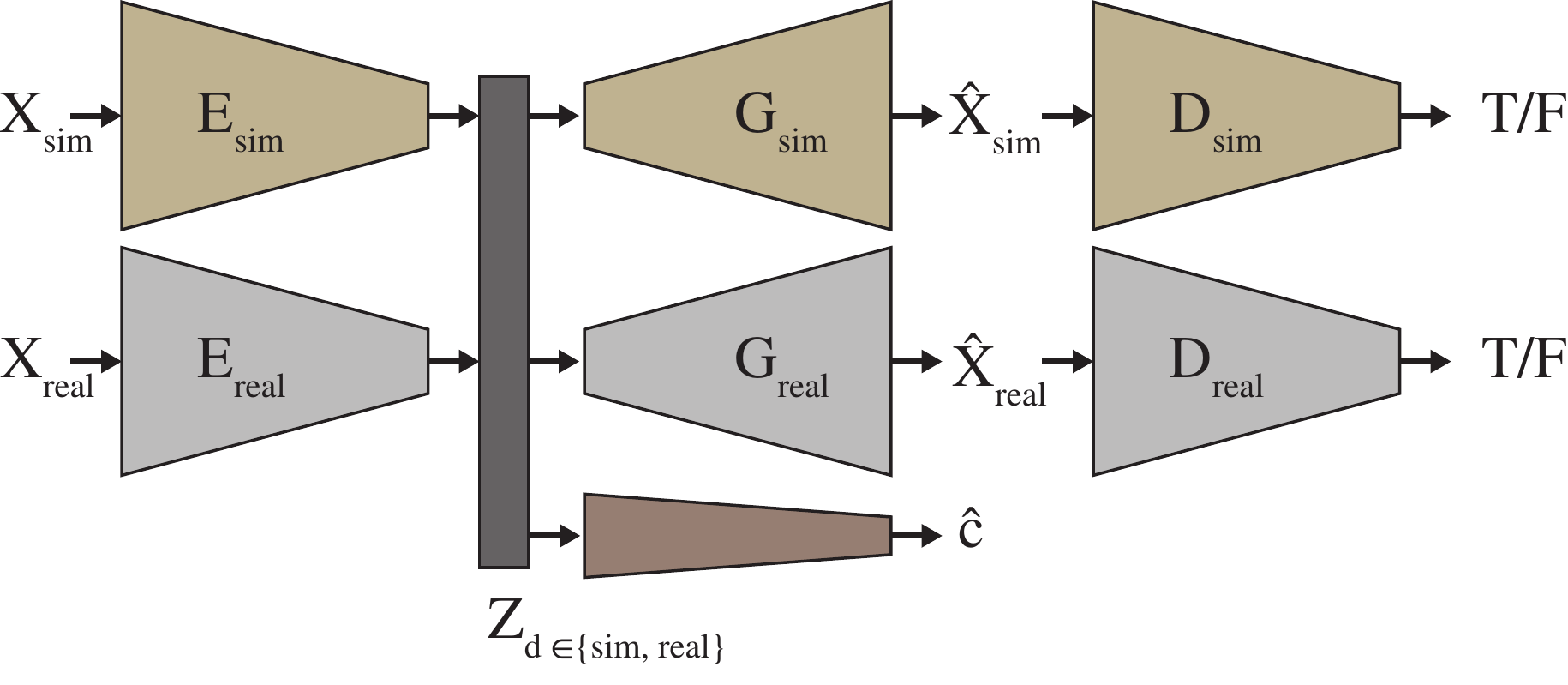}
\caption{Model architecture for domain-transfer from a simulated domain to real-world imagery, jointly learning control and domain translation. The encoders $E_{sim, real}$ map input images from their respective domains to a latent space $Z$ which is used for predicting vehicle controls $\hat{c}$. This common latent space is learned through direct and cyclic losses as part of learning image-to-image translation, indicated conceptually in Figure \ref{fig:losses} and in Section \ref{sec:losses}.}
\label{fig:model}
\end{figure}

Our framework consisted of various modules, which can broadly be broken down into an image-translator, a set of domain discriminator networks and the controller network for imitation learning. Fig. \ref{fig:model} provides a high-level overview of our model, showing the flow of information between the different modules.

\subsubsection{Translator}
The image-translation module followed the general framework proposed in \cite{Liu2017}. This consisted of two convolutional variational autoencoder-like networks where the latent embedding was swapped for translating between domains. More formally:
\begin{align}
\label{eq:domain2latent}
Z_{d} = & E_{d}(X_{d}) + \epsilon \\
\label{eq:latent2domain}
\hat{X}_{d} = & G_{d}(Z_d)
\end{align}
where $d \in [sim,real]$ represents the domain and $\epsilon \sim \mathit{N}(0,1) $ is noise added during training, but set to zero for inference. The process of translation consisted of computing $Z_d$ using one domain encoder and then predicting $\hat{X}$ with the other.

\subsubsection{Discriminators}
For each translated image a two-scale discriminator was used to align the appearance distribution of the translator output with the images in the corresponding domain. 

\subsubsection{Controllers}
The control architecture is broadly based on the end-to-end control architecture in \cite{Codevilla2018}, where the image translator forms the primary convolutional encoder. The latent tensor $Z_d$ is passed through CoordConv \cite{novotny2018semi} layers to learn a spatially aware embedding suited for control from this latent translation representation. Finally, the spatially aware tensor is reduced with global average pooling and then followed by fully connected layers.

\subsection{Losses}
\label{sec:losses}
Figure \ref{fig:losses} gives an overview of the main losses.

\subsubsection{Image Reconstruction Loss}
For a given domain $d$, $E_d$ and $G_d$ constitute a VAE. To improve image translation we used an L1 Loss $\mathcal{L}_{recon}$ between an image $X_d$ and the reconstructed image after passing it through the corresponding VAE, $X_d^{recon}~=~G_d(E_d(X_d))$, as shown in Figure \ref{fig:losses-recon}.

\subsubsection{Cyclic Reconstruction Loss}
Assuming a shared latent space implies the cycle-consistency constraint, which says that if an image is translated to the other domain and then translated back, the original image should be recovered \cite{CycleGAN2017,Liu2017}.
We applied a cyclic consistency loss $\mathcal{L}_{cyc}$ to the VAEs, given by an L1 loss between an image $X_d$ and the image after translating to the other domain, $d'$, and back, $X_d^{cyc}~=~G_d(E_{d'}(G_{d'}(E_d(X_d))))$, see Figure \ref{fig:losses-cycrecon}.

\subsubsection{Control Loss} 
\begin{figure}[t]
\centering
\vspace{-.4cm}
\subfloat[Reconstruction loss $\mathcal{L}_{recon}$\label{fig:losses-recon}]{\centering \includegraphics[width=0.48\columnwidth]{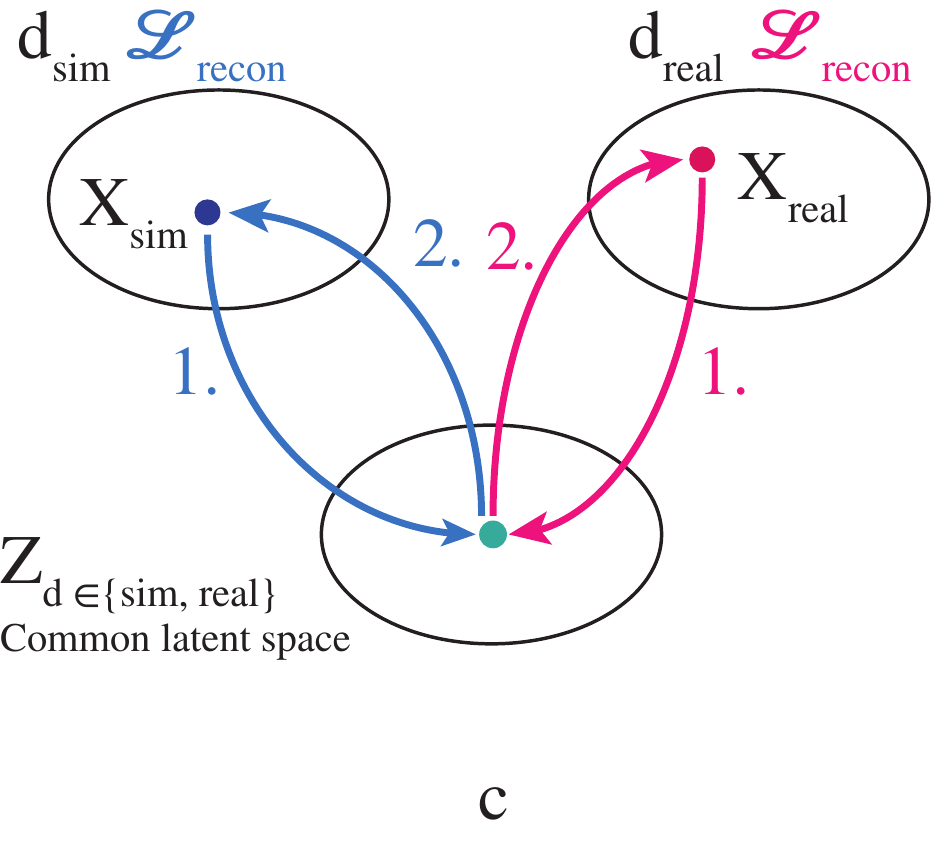}}
\subfloat[Cyclic reconstruction loss $\mathcal{L}_{cyc}$\label{fig:losses-cycrecon}]{\centering \includegraphics[width=0.48\columnwidth]
{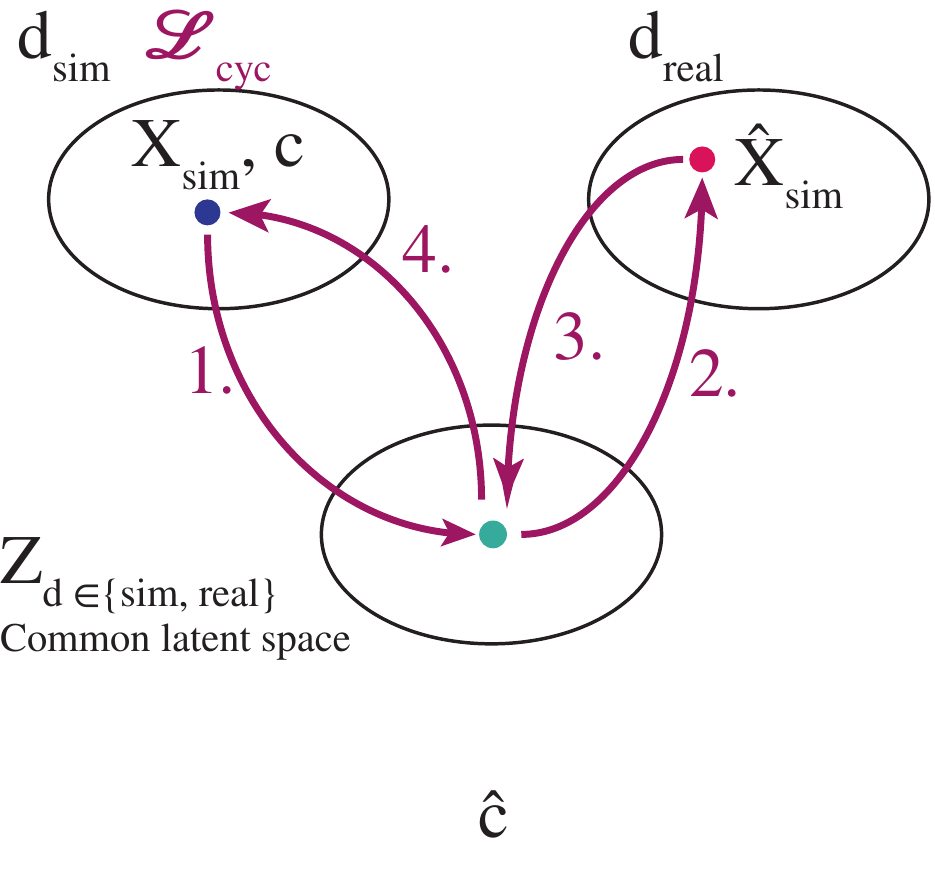}}
\hfill
\subfloat[Control loss $\mathcal{L}_{control}$\label{fig:losses-control}]{\centering \includegraphics[width=0.48\columnwidth]{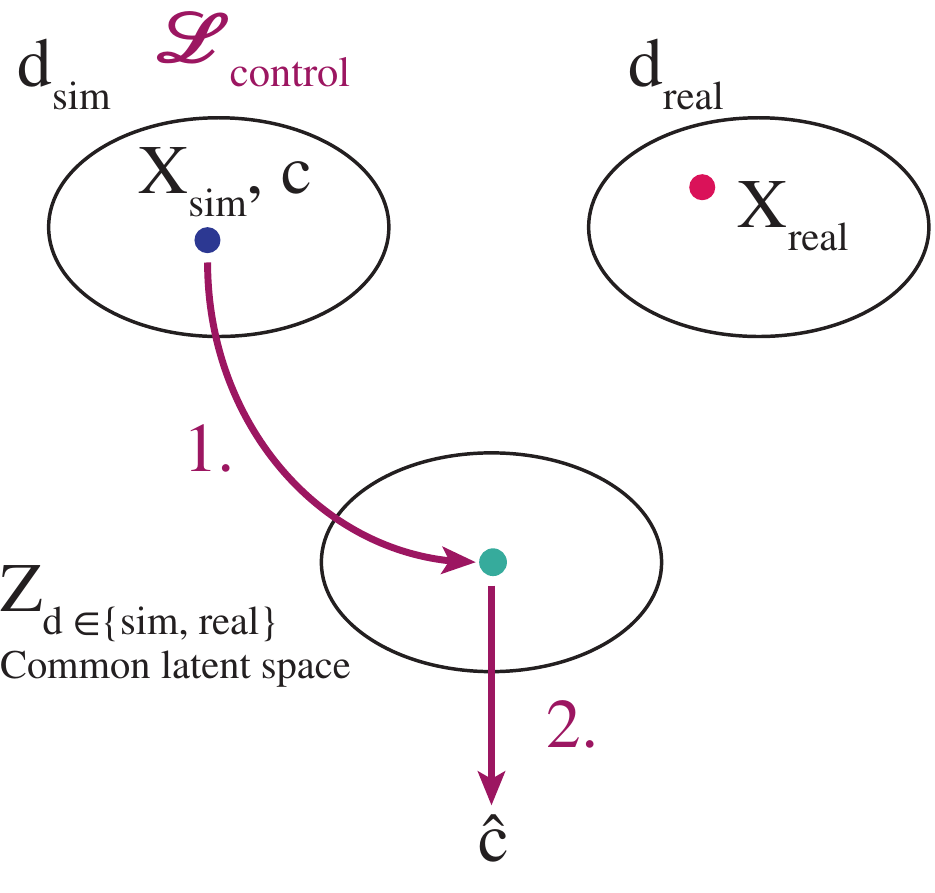}}
\subfloat[Cyclic control loss $\mathcal{L}_{cyc~control}$\label{fig:losses-cyccontrol}]{\centering \includegraphics[width=0.48\columnwidth]{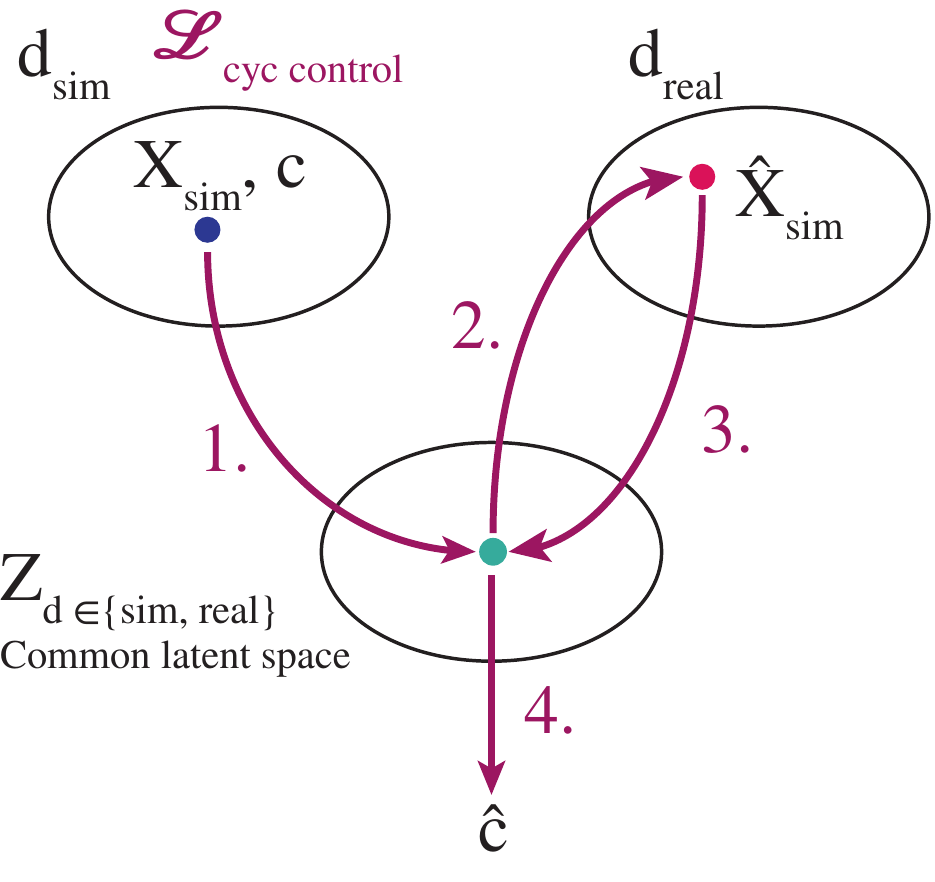}}
\caption{Learning the latent space $Z$ for translation and control requires a combination of direct and cyclic losses: learning to reconstruct images intra-domain, translating them inter-domain, while jointly learning a supervised control signal from the simulation domain. This figure does not illustrate the \textit{adversarial} or \textit{perceptual} loss. The \textit{latent reconstruction} loss is conceptually analogous to $\mathcal{L}_{cyc}$ in \ref{fig:losses-cycrecon}, but applied to the latent space. Section \ref{sec:losses} describes these loss terms in further detail. }
\label{fig:losses}
\end{figure}

To guide our model to learn features that are useful for driving, we also used a control loss $\mathcal{L}_{control}$, which is an L1 loss between the controller's predicted steering $\hat{c} = C(E_d(X_d))$ and the ground truth given by the autopilot, $c$, shown in Figure \ref{fig:losses-control}. \par
Control should be based on the semantic content of an image and independent of which domain the original image came from. We therefore introduced a cyclic control loss $\mathcal{L}_{cyc~control}$, as shown in Figure \ref{fig:losses-cyccontrol}, an L1 loss between the predicted steering $\hat{c}$ and the steering predicted by the controller when the image was translated to the other domain and then encoded, $\hat{c}^{cyc}~=~C(E_{d'}(G_{d'}(E_d(X_d))))$.

\subsubsection{Adversarial Loss}
Both the image-translator and the discriminators were optimised with the Least-Squares Generative Adversarial Network (LSGAN) objective proposed by \cite{mao2017least}. The discriminator \eqref{eq:dis-loss} and generator \eqref{eq:gen-loss} adversarial losses ensured that translated images resembled those from the chosen domain.
\begin{align}
\label{eq:dis-loss}
\mathcal{L}_{LSGAN}(D) = & \E_{X \sim p_{data}} [ (D(X) - 1)^2 ] + \\ 
& \E_{Z \sim p_Z(Z)} [(D(G(Z)) - 0)^2] \nonumber \\
\label{eq:gen-loss}
\mathcal{L}_{LSGAN}(G) = & \E_{Z \sim p_Z(Z)} [(D(G(Z)) - 1)^2]
\end{align}

\subsubsection{Perceptual Loss} To encourage consistent semantic content across the two domains, we employed the use of a pre-trained VGG \cite{Simonyan14c} model, which was applied to both the original and translated images. The perceptual loss $\mathcal{L}_{perceptual}$ was expressed as the difference between the features extracted from the last convolutional layer in the VGG16 model for a given input image and its translated counterpart. Extracted features were normalised via Instance Norm (IN) \cite{Ulyanov_2017_CVPR} following the result from Huang \etal \cite{Huang2018munit} demonstrating that applying IN before computing the feature distance makes the perceptual loss more domain-invariant.

\subsubsection{Latent Reconstruction Loss}
Ideally we wanted to encode the semantic content of the images within the latent space such that $Z$ is independent of the domain from which an image came. We therefore applied a latent reconstruction loss $\mathcal{L}_{Z recon}$, an L1 loss between the latent representation of an image $Z_d$ and the reconstruction of the latent representation after it was decoded to the other domain and then encoded once more, $Z_d^{recon} = E_{d'}(G_{d'}(Z_d))$. \par

The total discriminator loss for a given domain was $\mathcal{L}_{LSGAN}(D_d)$ \eqref{eq:dis-loss}. The VAEs and the controller were trained jointly using a weighted sum of losses with weights $\lambda_i$, given by
\begin{equation}
\label{equ:totloss}
\begin{aligned}[c]
\mathcal{L}_{tot} =& \lambda_{0} \mathcal{L}_{recon} + \lambda_{1} \mathcal{L}_{cyc} + \lambda_{2} \mathcal{L}_{control} \\
&+ \lambda_{3} \mathcal{L}_{cyc~control} + \lambda_{4} \mathcal{L}_{LSGAN}(G) \\
&+ \lambda_{5} \mathcal{L}_{perceptual} + \lambda_{6} \mathcal{L}_{Z recon}.
\end{aligned}
\end{equation}

The image translator was optimised using ADAM \cite{kingma2014adam}, with a learning rate of $0.0001$ and momentum terms of $0.5$ and $0.999$. The controller was trained using stochastic gradient descent with a learning rate of $0.01$.

\section{Evaluation}
\label{sec:evaluation}

To evaluate our approach in Section \ref{sec:method}, we considered data gathered from a simulated domain (with a procedurally generated environment) and a real domain (gathered from driving on real roads). We trained models using the methods in \ref{sec:transfer-learning-results} first in a rural setting, to compare our method to a number of baseline approaches. We then demonstrate our method scaling to the complexity of public urban roads in the United Kingdom.

Table \ref{table:open_loop} presents in-domain and cross domain open-loop (offline) metrics for the learned controller, and Table \ref{table:closed_loop} presents closed loop imitation control results evaluated on a physical vehicle in the rural environment. Qualitative results of the image-to-image translation are shown in Figure \ref{fig:translation} for both the rural and urban setting.
The following subsections detail the performance metrics used and test scenarios considered, followed by an analysis of the obtained results.

\subsection{Simulation Domain}
\subsubsection{Procedural Environment Generation}
\label{sec:proc-env-gen}
For the purpose of this work we created a simulated environment where we could build up a significant training set of image- and control-label pairs on procedurally generated virtual roads. Figure \ref{fig:car-view} illustrates the rural simulated world, while Figure \ref{fig:urban-setting} shows the urban simulated world.

For each data collection run in the rural world, a single road was created using a piece-wise B\'ezier curve generated by sampling 1-dimensional Simplex noise \cite{Perlin2002}. The curvature of the road could be coarsely controlled using the frequency parameter of the noise.  
The road surface was then assigned a random texture from a finite set of example road textures.
Once the rural road was constructed, trees and foliage were placed using Poisson Disk Sampling \cite{Bridson:2007:FPD:1278780.1278807} according to a foliage density parameter.  In simulation we were able to vary environmental factors, such as cloud cover, rainfall, surface water accumulation, and time of day. However, we fixed these as we wished to learn an image-to-action control policy in a very narrow source domain: we did not require high variance in the source distribution, rather to learn how the higher variance target distribution translates to the source.

For the urban world, we use a similar approach to procedurally generate a road network. We procedurally add buildings, trees and parked cars. Care was made to approximate the layout and topology of the real-world urban environment, but not the visual complexity or photo-realism. Instead, we rely on the image translation model to transfer the policy from the cartoon-like simulated world to the visually rich real-world.

\subsubsection{Simulated Expert Agent}
\label{sec:sim-expert-agent}

The labelled training set was generated from an expert autopilot agent. The expert driver has a simple proportional controller empirically tuned to track the lane based on ground-truth distance from the centre, maintaining a constant vehicle speed. 
The curve from Section \ref{sec:proc-env-gen} used to generate the road was used to generate a set of lane paths which were offset from the central curve. One end of the road was chosen at random and the simulated vehicle was placed at the end and told to follow the corresponding lane path.

To perturb the vehicle state, we used additive Ornstein-Uhlenbeck (OU) process noise \cite{uhlenbeck1930brownian} to the expert driver's actions. This had the effect of generating a more robust training dataset by moving the vehicle throughout the drivable lane space, observing the expert driver's response from each perturbed pose. The OU noise parameters $\theta$ and $\sigma$ were selected to maximise the magnitude of perturbation while still allowing the expert driver to largely stay within the lane.
This simulation environment ran asynchronously, with image-label pairs captured at 10Hz.

Leveraging this expert agent to learn to drive with imitation learning is only possible in simulation. It requires privileged information, such as the distance to the centre of the lane, which is not available in the real world. Furthermore, we perturb the state with noise to generate richer training data; this would be dangerous in the real world as it would require swerving on the road.

\subsection{Real-World Domain}
\subsubsection{Rural road}
\begin{figure}[tbp]
\centering
\subfloat[The 250m real-world rural driving route, coloured in blue.\label{fig:driving-route-real}]{\includegraphics[width=0.22\textwidth]{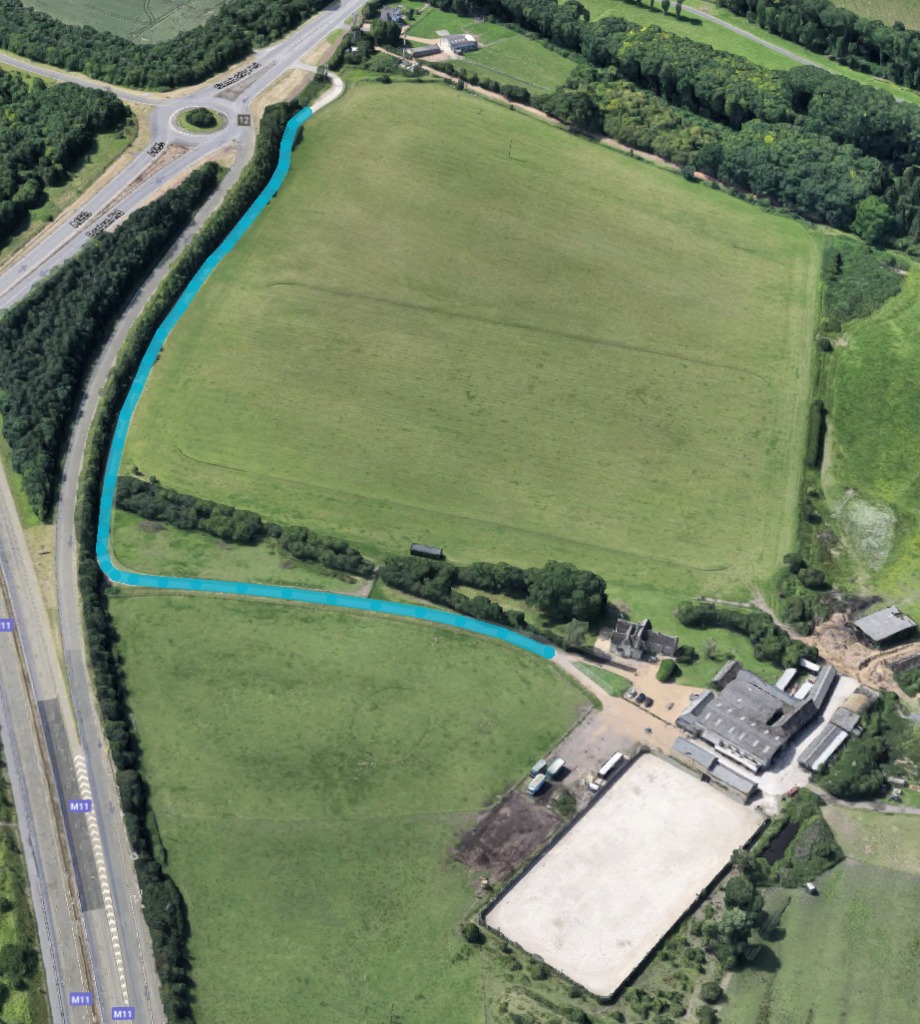}}\hfill
\subfloat[A procedurally generated curvy driving route in simulation.\label{fig:driving-route-sim}]{\includegraphics[width=0.22\textwidth]{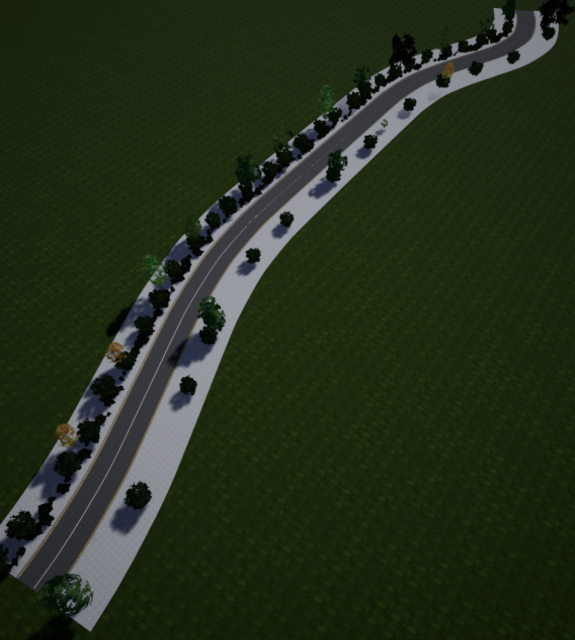}}
\\
\subfloat[Real-world rural road.\label{fig:1a}]{\includegraphics[width=0.22\textwidth]{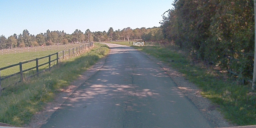}}\hfill
\subfloat[Simulated road.\label{fig:1b}]{\includegraphics[width=0.22\textwidth]{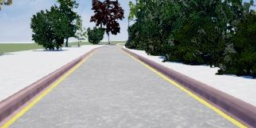}}
\caption[]{Rural setting: aerial views of the real-world (a) and simulated driving routes (b), along with example images from each domain showing the perspective from the vehicle (c, d).}
\label{fig:car-view}
\end{figure}

We considered a single driving environment for initial real-world testing, consisting of a 250m private one-lane rural road, shown in Figure \ref{fig:car-view}. For safety, the vehicle was only driven at $10~kmh^{-1}$ in the absence of any other vehicles wishing to use the road, both in data collection and under test.

\subsubsection{Urban road}
\begin{figure*}[tbp]
\centering
\subfloat[Simulated urban world.]{\includegraphics[height=3cm]{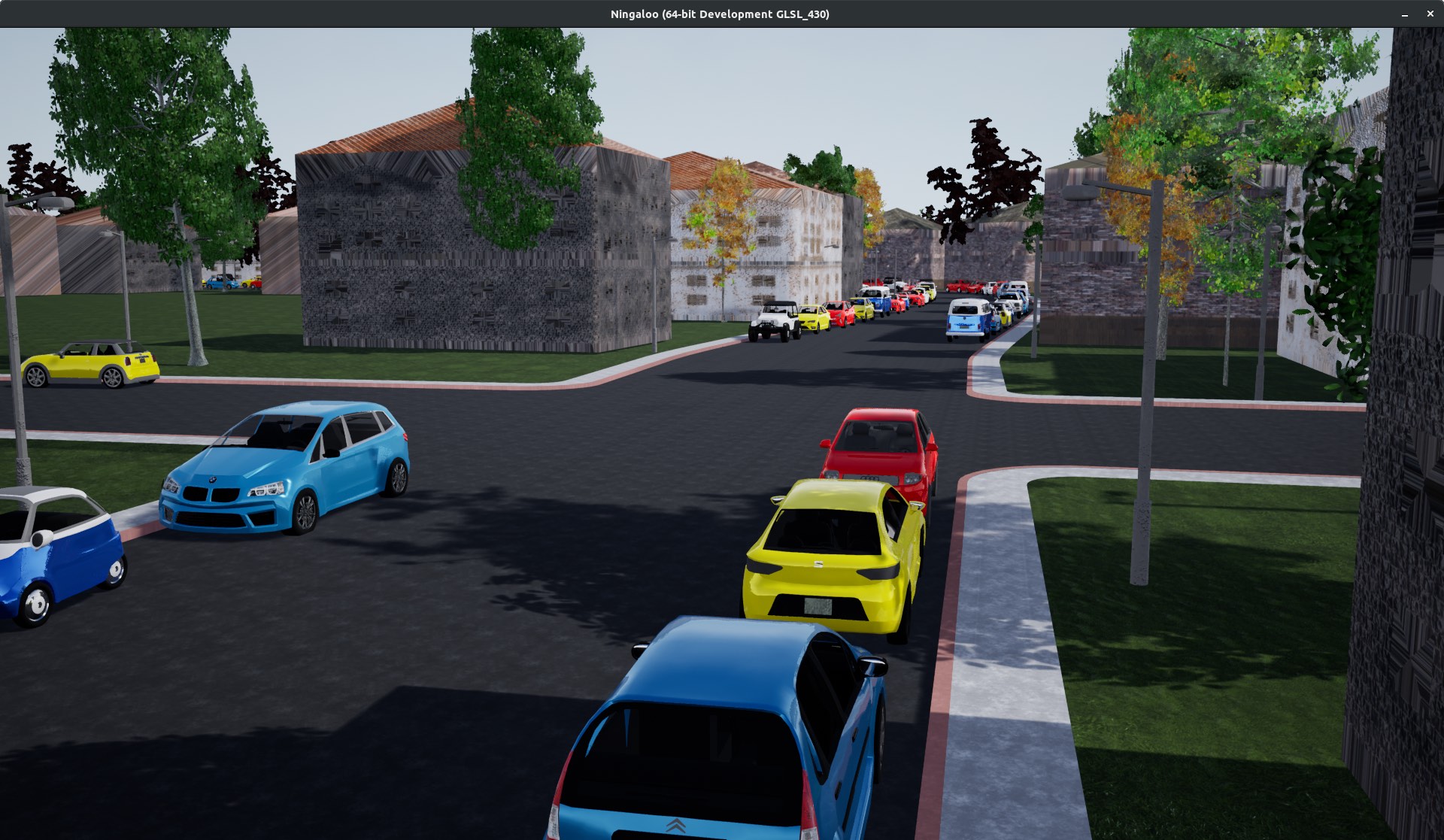}}\hfill
\subfloat[The real-world autonomous vehicle used for all experiments.]{\includegraphics[height=3cm]{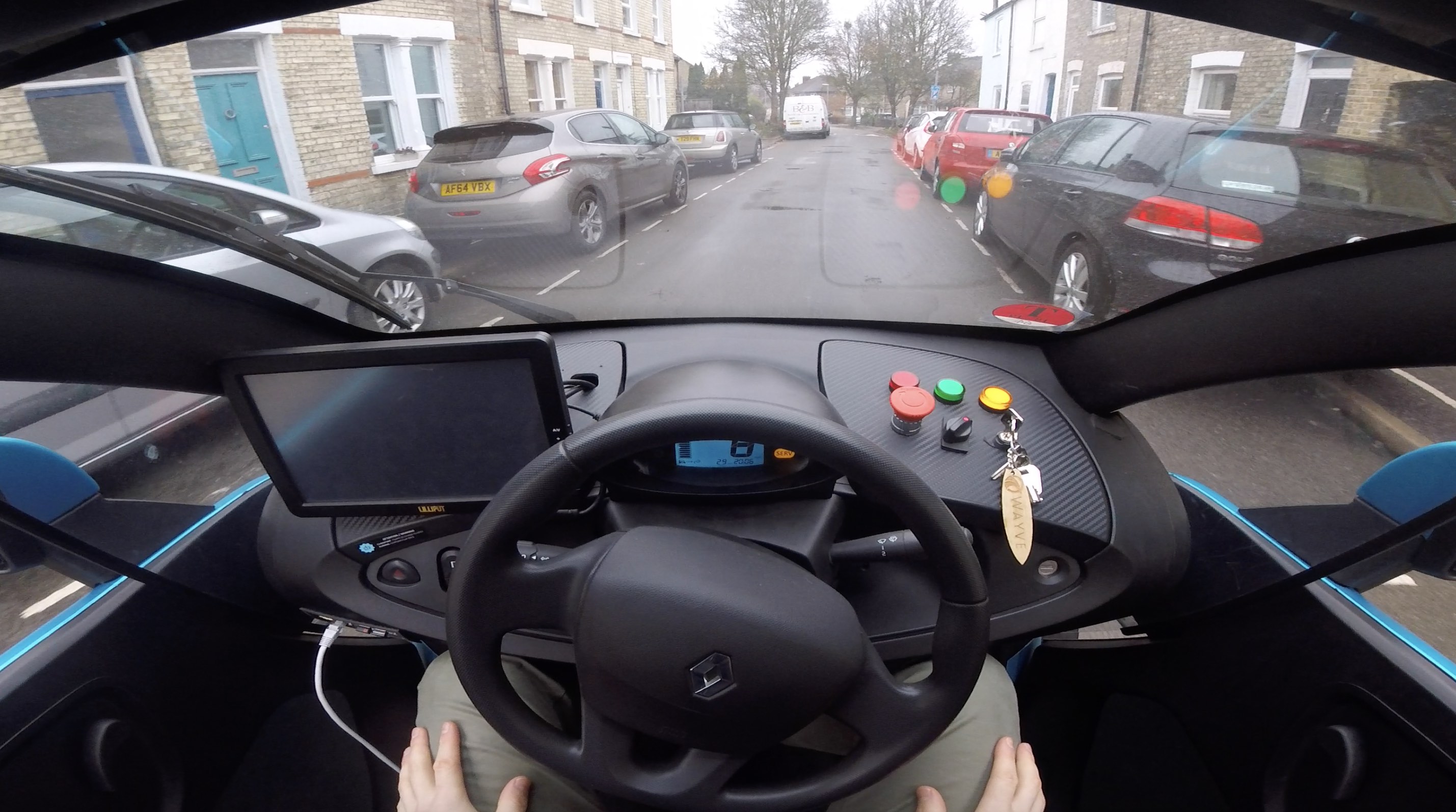}}\hfill
\subfloat[An example of the real-world urban roads used for closed loop testing in Cambridge, UK.]{\includegraphics[height=3cm]{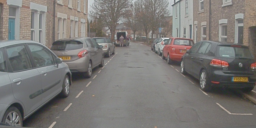}}
\caption[]{Urban setting: illustrations of the simulated and real-world domains for urban road scenes. Despite quite a large appearance change from the cartoon-like simulated world to the real-world, our model is able to successfully transfer the learned control policy to drive in in the real world, with no real world labels.}
\label{fig:urban-setting}
\end{figure*}

In addition, we consider an urban road environment in Cambridge, UK. We select minor public roads in dense suburban areas and opportunistically test when they are void of other traffic. Figure \ref{fig:urban-setting} illustrates typical scenes. We test across sun, overcast and raining weather conditions.

\subsection{Data}
\label{sec:data}
Table \ref{table:data} outlines the training, and test, lane following datasets used. For each domain, we collected approximately $60$k frames of training data and $20$k frames of test data.

\begin{table}[h]
	\caption{Datasets used for driving policy domain adaptation. Each simulation frame has associated expert control labels.}
	\label{table:data}
	\begin{center}
		\begin{tabular}{c c c}
			\toprule
			 & Training Frames & Test Frames \\
			\midrule
			 Simulation & 60014 & 17741 \\
			 Real & 57916 & 19141 \\
			\bottomrule
		\end{tabular}
	\end{center}
\end{table}

\subsubsection{Data Balancing}
\label{sec:data-balancing}

\begin{figure}
\centering
\includegraphics[width=0.8\columnwidth]{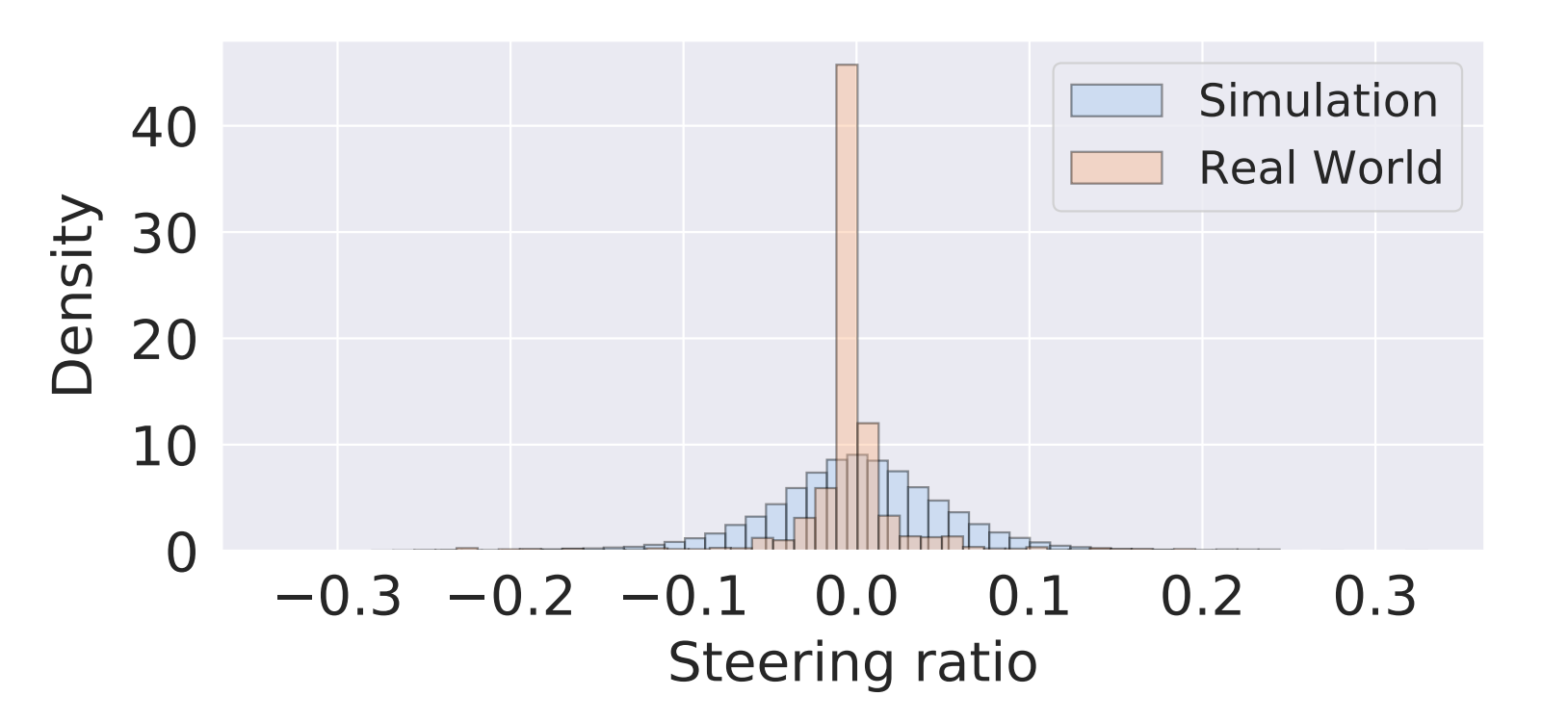}
\caption[]{The steering data distribution (in the range $\pm1.0$) differs dramatically between the simulation and real world, where simulation has much greater variance. The real-world data has shorter tails and is heavily concentrated around the zero position, due to the profile of the rural driving route.}
\label{fig:steering-distribution}
\end{figure}

The maximum steering range of each vehicle was parameterised to $\pm1.0$. Due to the nature of driving, the labelled training set was heavily dominated by data with near-zero steering angles (i.e., driving straight), as shown in Figure \ref{fig:steering-distribution}. If the controller is trained with a naive mean absolute error (MAE) loss, it could feasibly predict zero steering for all outputs and fail to learn the correct image-steering mapping.
 
To address this problem, we split the data into eight bins according to the steering angle, and uniformly sampled from each bin, upsampling the data such that every bin contained the same number of samples as the bin with the maximum number of data. The bins edges used in our experiments were $[-1, -0.075, -0.05, -0.025, 0.0, 0.025, 0.05, 0.075, 1]$. The steering limits differ between vehicles, hence we applied a linear calibration to the real vehicle steering output.

\subsection{Transfer Learning Results}
\label{sec:transfer-learning-results}

\subsubsection{Baselines}
We compared our method to the following baselines:
\begin{itemize}
	\item \textbf{Simple Transfer} this takes a model pre-trained in simulation and directly applies it to the real-world data. Note: compared to the following this model only sees examples from simulation.
	\item \textbf{Real-to-Sim Translation} uses the unsupervised image-to-image translation to convert a real-world image to the simulation domain for direct application of the controller pre-trained in simulation.
	\item \textbf{Sim-to-Real Translation} uses the unsupervised image-to-image translation for training a controller on translated images (sim-to-real) with corresponding simulated actions. At test time no translation was performed and only the controller was used.
	\item \textbf{Latent Feature ADA} applies Adversarial Domain Adaptation (ADA) \cite{wulfmeier2017addressing} to the feature space to align encoders from simulation and real data. 
\end{itemize}
For evaluation, we also compared a Drive-Straight policy as a proxy to assess road curvature, and to quantitatively assess the efficacy of offline metrics.

\subsubsection{Performance Metrics}
\label{sec:metrics}
Open-loop evaluation of control policies is an open question: \cite{Codevilla_2018_ECCV} and \cite{Yang2018} present a range of metrics, demonstrating weak correlation between offline metrics and online performance. We adopted two metrics to compare driving policies: mean average error (MAE), and Balanced MAE. These metrics for all approaches tested here are outlined in Table \ref{table:open_loop}. The MAE between the predicted and ground-truth steering is a useful loss function, but poor reflection of actual performance due to the data imbalance. We propose a Balanced-MAE metric: discretising the validation data as per the bins in Section \ref{sec:data-balancing}, computing the MAE of the data per bin, then taking an equally weighted average across all bins (similar in principle to mean AP in object detection or recognition tasks). Qualitatively, we found that this metric correlates with closed-loop driving performance in simulation.

For closed-loop testing on our rural driving route, we used a simple measure of distance travelled per safety driver intervention in metres travelled, averaged over 3km of driving. During test, we ran the models at close to the camera's 30Hz frame rate. Table \ref{table:closed_loop} outlines the performance of our system, as well as a number of baseline models. We found that the open-loop metrics in Table \ref{table:open_loop} hold only a weak correlation with real-world performance, demonstrated by the difference in closed- and open-loop performance between our model and the baseline approaches. Clearly, further work is needed to develop open-loop metrics which support real-world performance.

\begin{table}[h]
	\caption{Open-loop control metrics on the simulation and real test datasets from Table \ref{table:data}.}
	\label{table:open_loop}
	\begin{center}
		\begin{tabular}{c c c c c}
        	\toprule
			& \multicolumn{2}{c}{Simulation} & \multicolumn{2}{c}{Real} \\
            \midrule
			 & MAE & Bal-MAE & MAE & Bal-MAE \\
			\midrule
             Drive-Straight & 0.043 & 0.087 & \textbf{0.019} & 0.093 \\
			 Simple Transfer & 0.05 & 0.055 & 0.265 & 0.272 \\
			 Real-to-Sim Translation & - & - & 0.261 & 0.234\\
			 Sim-to-Real Translation & - & - & 0.059 & \textbf{0.045}\\
			 Latent Feature ADA & 0.040 & 0.047 & 0.032 & 0.071\\ 
             \midrule
			 \textbf{Ours} & \textbf{0.017} & \textbf{0.018} & 0.081 & 0.087\\
			\bottomrule
		\end{tabular}
	\end{center}
\end{table}

\vspace{-0.15cm}
\begin{table}[h]
	\caption{On-vehicle performance, driving 3km on the rural driving route in \ref{fig:driving-route-real}. For policies unable to drive a lap with $\leq1$ intervention, we terminated after one $250$m lap (\dag).}
	\label{table:closed_loop}
	\begin{center}
		\begin{tabular}{c c}
			\toprule
			 & Mean distance / intervention (metres) \\
			\midrule
			Drive-straight & $23^{\dag}$ \\
            Simple Transfer & $9^{\dag}$ \\
            Real-to-Sim Translation & $10^{\dag}$ \\
            Sim-to-Real Translation & $28^{\dag}$ \\
            Latent Feature ADA & $15^{\dag}$ \\
            \midrule
			\textbf{Ours} & No intervention over 3km  \\
			\bottomrule
		\end{tabular}
	\end{center}
\end{table}

\subsubsection{Loss Ablation}
The various terms in Eqn. \ref{equ:totloss} are designed to work in concert to maintain a structured latent space to facilitate zero-shot transfer from simulated label to the real domain. The following ablation study investigates the influence of each term by setting the corresponding weighting $\lambda$ to zero, effectively removing it from the training procedure. Table \ref{table:ablation} shows that all terms play a significant role in reducing the error with the exception of the cyclic control factor. This indicated that the other cyclic losses are sufficient to maintain a shared latent structure. Not surprisingly, the L1 cyclic reconstruction and GAN loss from \cite{Liu2017} are most critical for control transfer. Finally, as removing the control term prevented the network from learning anything sensible, we instead evaluated the effect of the gradients of the controller on the image translator. Here we zeroed the gradients from the controller before flowing back into the translator, effectively turning the controller into a passive neural stethoscope, as described in \cite{fuchs2018neural}. These gradients would normally help inject structure for the task into the translator in an auxiliary fashion leading to better real-world performance. This increase in error by not optimising the encoders for both tasks could also be a contributing factor to the poor performance of the translation baselines (real-to-sim and sim-to-real).

\begin{table}[h]
	\caption{Ablation by removing single terms in eqn. \ref{equ:totloss}. Values expressed as the average error multiplier across three runs for the open-loop metrics when transferred to the real domain test dataset. For the Control term the gradients applied to the image translator for this loss are removed (\ddag).}
	\label{table:ablation}
	\begin{center}
		\begin{tabular}{c c c}
        	\toprule
			 & MAE & Bal-MAE  \\
			\midrule
             Recon. L1 & 1.602 & 1.259\\
			 Recon. Cyclic L1 & 2.980 & 2.395 \\
			 $Z$ Recon. & 1.766 & 1.270 \\
			 Perceptual & 1.693 & 1.435 \\
			 Cyc Control & 0.923 & 1.068 \\
			 GAN & 1.893 & 1.507 \\ 
			 Control-grad(\ddag) & 1.516 & 1.433 \\ 
			\bottomrule
		\end{tabular}
	\end{center}
\end{table}

\begin{figure*}[h]
\centering
\subfloat[Rural environment.\label{fig:translation-rural}]{\includegraphics[width=0.48\textwidth]{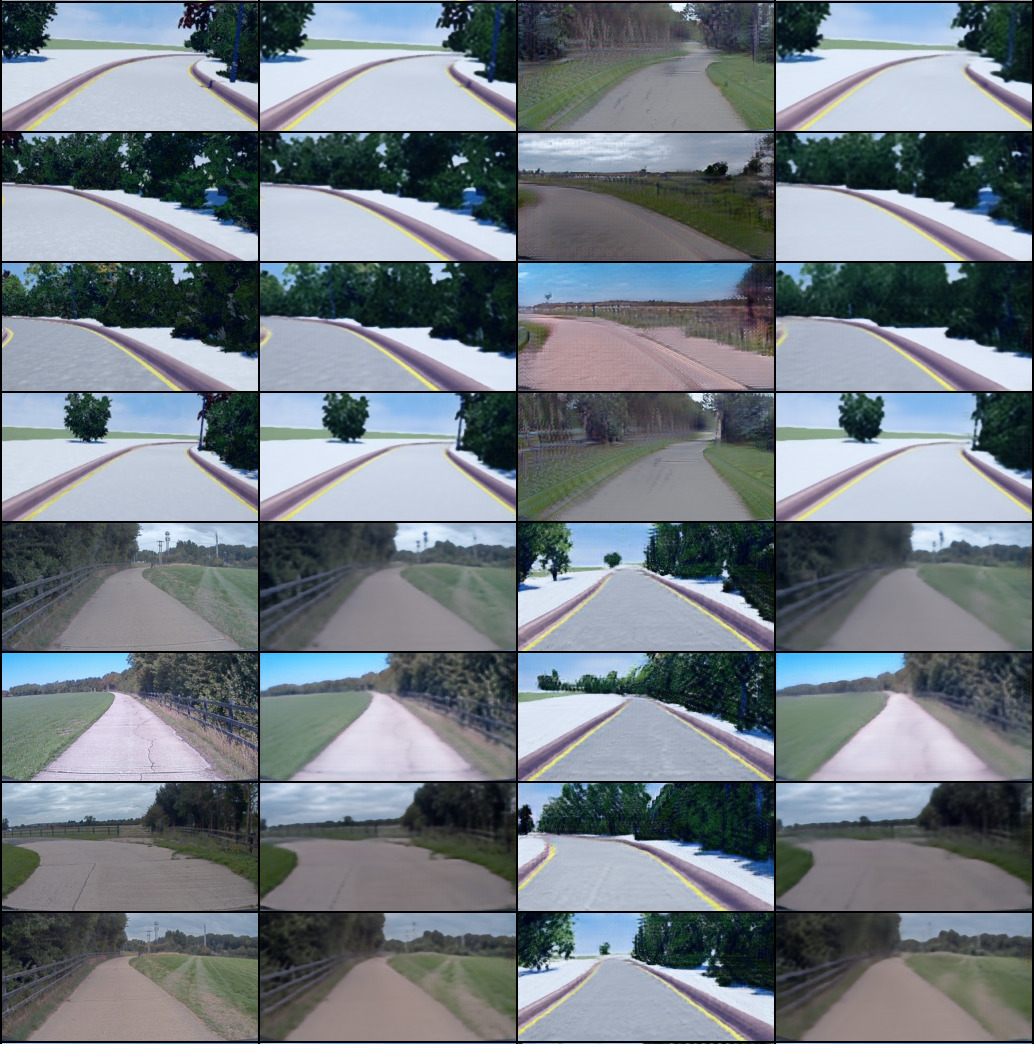}}\hfill
\subfloat[Urban environment.\label{fig:translation-urban}]{\includegraphics[width=0.48\textwidth]{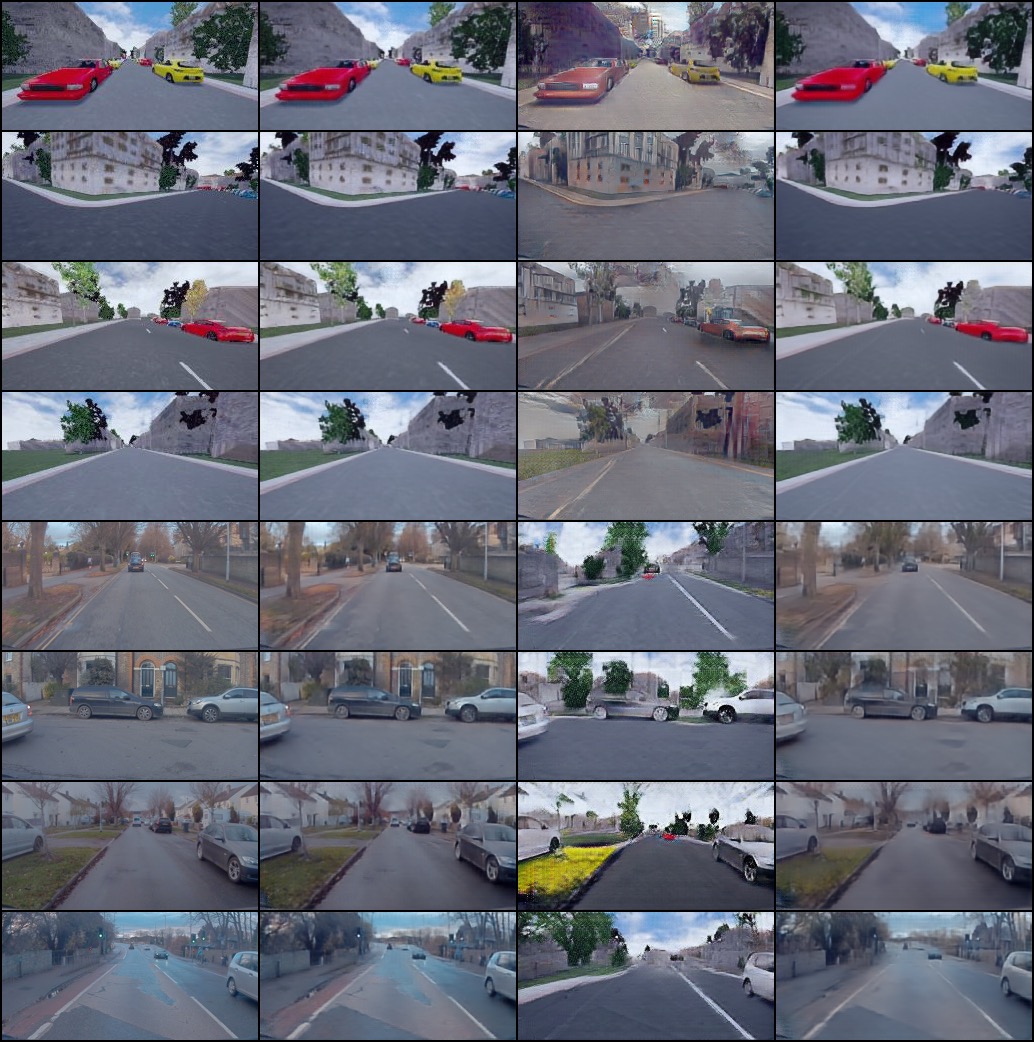}}
\caption{Qualitative results of the image translation model for (a) rural and (b) urban environments. In each example, the top four rows show images originating in our simulated environment, and the bottom half showing translation from real world images. The columns left to right are as follows: original, image reconstructed from latent back to original domain, translated to other domain, and taking the translated image and translating it back to the original domain.}
\label{fig:translation}
\vspace{-4mm}
\end{figure*}

\subsubsection{Urban Road Results}
To further demonstrate the efficacy of our method, we performed closed-loop testing on public UK urban roads. Using our model, we were able to successfully demonstrate the vehicle lane-following on dense urban side streets in Cambridge, UK. Examples are shown in Figure \ref{fig:translation-urban}.

We observed one main conclusion: the model is typically able to drive if, and only if, the visual translation is successful. For example, when the topology of the road and all cars are translated correctly between domains, the car's control policy gives the desired lane-following closed-loop behaviour. When a car is mis-translated into a footpath, the control policy requires intervention. Therefore, we conclude that this method may be able to scale to more complex domains, beyond simple lane following tasks, if the simulator is capable of simulating these scenarios and we can learn a successful domain translation model.

\subsubsection{Visualisation}
Understanding the performance of a control policy coupled with image domain transfer can be difficult. 
The bi-directional image translator provides the ability to inspect the model's interpretation of the road surface as shown in Fig. \ref{fig:translation}. Here we can observe that the curvature and offset of the road is appropriately translated across domains facilitating a consistent transfer of the steering signal. Interestingly, in the third row of Figure \ref{fig:translation-rural} (where the vehicle is nearly driving off the virtual road, which is far from the distribution of collected real images) we notice the model will generate an image closer to the distribution of real-world observations. In such scenarios, the controller is trained with a high magnitude steering command to correct its course in simulation, which results in robust behaviour in the real world as it controlled the vehicle back into the distribution of real observations.
 
	
\section{Conclusions}

Learning a driving policy from simulation has many advantages: training data is cheap, auxiliary ground-truth information can be provided with ease, and the vehicle can be put in situations that are difficult or dangerous to undertake in reality. Previously, with the substantial gap in complexity between the two domains, it was considered infeasible to transfer driving policies from simulation to the real world.

In this work, we present the first system that is capable of leveraging simulation to learn an end-to-end driving policy to directly transfer to real-world scenarios without any additional human demonstrations. This model jointly learns to translate images between the real-world operating domain and a procedurally generated simulation environment, while also learning to predict control decisions off of the latent space between the two domains given only the ground-truth labels from simulation. We empirically validated our proposed model in closed-loop against several baselines, successfully driving 3km between interventions on a real-world lane following task. We further evaluated the model using several standard open-loop metrics, observing that these metrics ultimately did not prove predictive of driving performance. Finally, we demonstrated this system driving in closed-loop on public urban roads in the United Kingdom.

This work provides evidence that end-to-end policy learning and simulation-to-reality transfer are highly promising directions for the development of autonomous driving systems. We note that standard open-loop metrics for this problem need to be improved, and leave this question to future work. Furthermore, the addition of orthogonal, but relevant, approaches on temporal motion consistency \cite{wang2018vid2vid, zhang2018vr} could further advance this field. We hope this work inspires further investigation of both learning driving policies directly from data, and exploiting simulation for removing the constraints of the real world.

\addtolength{\textheight}{-2cm}   




\newpage
\bibliographystyle{plain}
\bibliography{references/sim2real}  

\begin{thebibliography}{10}

\bibitem{bojarski2016end}
Mariusz Bojarski, Davide Del~Testa, Daniel Dworakowski, Bernhard Firner, Beat
  Flepp, Prasoon Goyal, Lawrence~D Jackel, Mathew Monfort, Urs Muller, Jiakai
  Zhang, et~al.
\newblock End to end learning for self-driving cars.
\newblock {\em arXiv preprint arXiv:1604.07316}, 2016.

\bibitem{Bridson:2007:FPD:1278780.1278807}
Robert Bridson.
\newblock Fast poisson disk sampling in arbitrary dimensions.
\newblock In {\em ACM SIGGRAPH 2007 Sketches}, SIGGRAPH '07, New York, NY, USA,
  2007. ACM.

\bibitem{Codevilla_2018_ECCV}
Felipe Codevilla, Antonio~M. Lopez, Vladlen Koltun, and Alexey Dosovitskiy.
\newblock On offline evaluation of vision-based driving models.
\newblock In {\em The European Conference on Computer Vision (ECCV)}, September
  2018.

\bibitem{Codevilla2018}
Felipe Codevilla, Matthias M{\"u}ller, Antonio L{\'o}pez, Vladlen Koltun, and
  Alexey Dosovitskiy.
\newblock End-to-end driving via conditional imitation learning.
\newblock In {\em International Conference on Robotics and Automation (ICRA)},
  2018.

\bibitem{Dosovitskiy17}
Alexey Dosovitskiy, German Ros, Felipe Codevilla, Antonio Lopez, and Vladlen
  Koltun.
\newblock {CARLA}: {An} open urban driving simulator.
\newblock In {\em Proceedings of the 1st Annual Conference on Robot Learning},
  pages 1--16, 2017.

\bibitem{fuchs2018neural}
Fabian~B Fuchs, Oliver Groth, Adam~R Kosoriek, Alex Bewley, Markus Wulfmeier,
  Andrea Vedaldi, and Ingmar Posner.
\newblock Neural stethoscopes: Unifying analytic, auxiliary and adversarial
  network probing.
\newblock {\em arXiv preprint arXiv:1806.05502}, 2018.

\bibitem{gaidon2016virtual}
Adrien Gaidon, Qiao Wang, Yohann Cabon, and Eleonora Vig.
\newblock Virtual worlds as proxy for multi-object tracking analysis.
\newblock In {\em Proceedings of the IEEE conference on computer vision and
  pattern recognition}, pages 4340--4349, 2016.

\bibitem{Hoffman2018cycada}
Judy Hoffman, Eric Tzeng, Taesung Park, Jun-Yan Zhu, Phillip~Isola Openai, Kate
  Saenko, Alexei~A Efros, and Trevor Darrell.
\newblock {CyCADA: Cycle-Consistent Adversarial Domain Adaptation}.
\newblock In {\em International Conference on Machine Learning (ICML)}, 2018.

\bibitem{hong2018virtual}
Zhang-Wei Hong, Chen Yu-Ming, Shih-Yang Su, Tzu-Yun Shann, Yi-Hsiang Chang,
  Hsuan-Kung Yang, Brian Hsi-Lin Ho, Chih-Chieh Tu, Yueh-Chuan Chang, Tsu-Ching
  Hsiao, et~al.
\newblock Virtual-to-real: Learning to control in visual semantic segmentation.
\newblock In {\em International Joint Conference on Artificial Intelligence
  ({IJCAI})}, 2018.

\bibitem{Huang2018munit}
Xun Huang, Ming-Yu Liu, Serge Belongie, and Jan Kautz.
\newblock {Multimodal Unsupervised Image-to-Image Translation}.
\newblock In {\em European Conference on Computer Vision (ECCV)}, 2018.

\bibitem{Isola_2017_CVPR}
Phillip Isola, Jun-Yan Zhu, Tinghui Zhou, and Alexei~A. Efros.
\newblock Image-to-image translation with conditional adversarial networks.
\newblock In {\em The IEEE Conference on Computer Vision and Pattern
  Recognition (CVPR)}, July 2017.

\bibitem{kendall2018learning}
Alex Kendall, Jeffrey Hawke, David Janz, Przemyslaw Mazur, Daniele Reda,
  John-Mark Allen, Vinh-Dieu Lam, Alex Bewley, and Amar Shah.
\newblock Learning to drive in a day.
\newblock {\em arXiv preprint arXiv:1807.00412}, 2018.

\bibitem{kingma2014adam}
Diederik~P. Kingma and Jimmy Ba.
\newblock Adam: {A} method for stochastic optimization.
\newblock {\em CoRR}, abs/1412.6980, 2014.

\bibitem{li2016playing}
Li~Li, Rongrong Chen, and Jing Chen.
\newblock Playing action video games improves visuomotor control.
\newblock {\em Psychological science}, 27(8):1092--1108, 2016.

\bibitem{Liu2017}
Ming-Yu Liu, Thomas Breuel, and Jan Kautz.
\newblock {Unsupervised Image-to-Image Translation Networks}.
\newblock In {\em Advances in Neural Information Processing Systems (NIPS)},
  2017.

\bibitem{Liu2016a}
Ming-Yu Liu and Oncel Tuzel.
\newblock {Coupled Generative Adversarial Networks}.
\newblock In {\em Advances in Neural Information Processing Systems}, 2016.

\bibitem{mao2017least}
Xudong Mao, Qing Li, Haoran Xie, Raymond~YK Lau, Zhen Wang, and Stephen~Paul
  Smolley.
\newblock Least squares generative adversarial networks.
\newblock In {\em Computer Vision (ICCV), 2017 IEEE International Conference
  on}, pages 2813--2821. IEEE, 2017.

\bibitem{mehta2018learning}
Ashish Mehta, Adithya Subramanian, and Anbumani Subramanian.
\newblock Learning end-to-end autonomous driving using guided auxiliary
  supervision.
\newblock {\em arXiv preprint arXiv:1808.10393}, 2018.

\bibitem{muller2018driving}
Matthias M{\"u}ller, Alexey Dosovitskiy, Bernard Ghanem, and Vladen Koltun.
\newblock Driving policy transfer via modularity and abstraction.
\newblock {\em arXiv preprint arXiv:1804.09364}, 2018.

\bibitem{muller2006off}
Urs Muller, Jan Ben, Eric Cosatto, Beat Flepp, and Yann~L Cun.
\newblock Off-road obstacle avoidance through end-to-end learning.
\newblock In {\em Advances in neural information processing systems}, pages
  739--746, 2006.

\bibitem{novotny2018semi}
David Novotny, Samuel Albanie, Diane Larlus, and Andrea Vedaldi.
\newblock Semi-convolutional operators for instance segmentation.
\newblock {\em arXiv preprint arXiv:1807.10712}, 2018.

\bibitem{Perlin2002}
Ken Perlin.
\newblock Improving noise.
\newblock {\em ACM Trans. Graph.}, 21(3):681--682, July 2002.

\bibitem{pomerleau1989alvinn}
Dean~A Pomerleau.
\newblock Alvinn: An autonomous land vehicle in a neural network.
\newblock In {\em Advances in neural information processing systems}, pages
  305--313, 1989.

\bibitem{richter2017playing}
Stephan~R Richter, Zeeshan Hayder, and Vladlen Koltun.
\newblock Playing for benchmarks.
\newblock In {\em International conference on computer vision (ICCV)},
  volume~2, 2017.

\bibitem{richter2016playing}
Stephan~R Richter, Vibhav Vineet, Stefan Roth, and Vladlen Koltun.
\newblock Playing for data: Ground truth from computer games.
\newblock In {\em European Conference on Computer Vision}, pages 102--118.
  Springer, 2016.

\bibitem{ros2016synthia}
German Ros, Laura Sellart, Joanna Materzynska, David Vazquez, and Antonio~M
  Lopez.
\newblock The synthia dataset: A large collection of synthetic images for
  semantic segmentation of urban scenes.
\newblock In {\em Proceedings of the IEEE conference on computer vision and
  pattern recognition}, pages 3234--3243, 2016.

\bibitem{Simonyan14c}
K.~Simonyan and A.~Zisserman.
\newblock Very deep convolutional networks for large-scale image recognition.
\newblock {\em CoRR}, abs/1409.1556, 2014.

\bibitem{Tobin2017}
Josh Tobin, Rachel Fong, Alex Ray, Jonas Schneider, Wojciech Zaremba, and
  Pieter Abbeel.
\newblock {Domain Randomization for Transferring Deep Neural Networks from
  Simulation to the Real World}.
\newblock In {\em IEEE/RSJ International Conference on Intelligent Robots and
  Systems (IROS)}, 2017.

\bibitem{tzeng2016adapting}
Eric Tzeng, Coline Devin, Judy Hoffman, Chelsea Finn, Pieter Abbeel, Sergey
  Levine, Kate Saenko, and Trevor Darrell.
\newblock Adapting deep visuomotor representations with weak pairwise
  constraints.
\newblock In {\em Workshop on Algorithmic Foundations in Robotics (WAFR)},
  2016.

\bibitem{uhlenbeck1930brownian}
G.~E. Uhlenbeck and L.~S. Ornstein.
\newblock On the theory of the brownian motion.
\newblock {\em Phys. Rev.}, 36:823--841, Sep 1930.

\bibitem{Ulyanov_2017_CVPR}
Dmitry Ulyanov, Andrea Vedaldi, and Victor Lempitsky.
\newblock Improved texture networks: Maximizing quality and diversity in
  feed-forward stylization and texture synthesis.
\newblock In {\em The IEEE Conference on Computer Vision and Pattern
  Recognition (CVPR)}, July 2017.

\bibitem{wang2018vid2vid}
Ting-Chun Wang, Ming-Yu Liu, Jun-Yan Zhu, Guilin Liu, Andrew Tao, Jan Kautz,
  and Bryan Catanzaro.
\newblock Video-to-video synthesis.
\newblock In {\em Advances in Neural Information Processing Systems (NeurIPS)},
  2018.

\bibitem{wenzel2018modular}
Patrick Wenzel, Qadeer Khan, Daniel Cremers, and Laura Leal-Taixe.
\newblock Modular vehicle control for transferring semantic information between
  weather conditions using gans.
\newblock In {\em Conference on Robot Learning}, pages 253--269, 2018.

\bibitem{wulfmeier2017addressing}
Markus Wulfmeier, Alex Bewley, and Ingmar Posner.
\newblock {Addressing appearance change in outdoor robotics with adversarial
  domain adaptation}.
\newblock In {\em IEEE International Conference on Intelligent Robots and
  Systems (IROS)}, 2017.

\bibitem{wulfmeier2017incremental}
Markus Wulfmeier, Alex Bewley, and Ingmar Posner.
\newblock Incremental adversarial domain adaptation for continually changing
  environments.
\newblock In {\em International Conference on Robotics and Automation (ICRA)},
  2018.

\bibitem{Yang2018}
Luona Yang, Xiaodan Liang, Tairui Wang, and Eric Xing.
\newblock {Real-to-Virtual Domain Unification for End-to-End Autonomous
  Driving}.
\newblock Technical report.

\bibitem{zhang2018vr}
Jingwei Zhang, Lei Tai, Yufeng Xiong, Ming Liu, Joschka Boedecker, and Wolfram
  Burgard.
\newblock Vr goggles for robots: Real-to-sim domain adaptation for visual
  control.
\newblock {\em arXiv preprint arXiv:1802.00265}, 2018.

\bibitem{CycleGAN2017}
Jun-Yan Zhu, Taesung Park, Phillip Isola, and Alexei~A Efros.
\newblock Unpaired image-to-image translation using cycle-consistent
  adversarial networks.
\newblock In {\em Computer Vision (ICCV), 2017 IEEE International Conference
  on}, 2017.

\end{thebibliography}

\end{document}